\documentclass[runningheads]{llncs}
\usepackage[T1]{fontenc}
\usepackage{graphicx}
\usepackage{booktabs}
\usepackage[misc]{ifsym}

\usepackage{url}
\usepackage{hyperref}
\usepackage{amsmath}
\usepackage{amssymb}
\usepackage{placeins}

\usepackage{todonotes}

\begin{document}

\title{Equivariant Flow Matching for Symmetry-Breaking Bifurcation Problems}
\titlerunning{Equivariant Flow Matching for Symmetry-Breaking Bifurcation Problems}

% % \author{Author information scrubbed for double-blind reviewing}
% \author{Andr\'e Lauren Benjamin\inst{1} \and
% Calvin Cordozar Broadus Jr.\inst{2,3} \corr \and
% Antwan Andr\'e Patton\inst{1}\orcidID{0000-1111-2222-3333}}

\author{Fleur Hendriks\inst{1,2}%\orcidID{0000-0001-6048-7357}
\and
  Ond\v{r}ej Roko\v{s}\inst{1}%\orcidID{0000-0003-2589-5333}
  \and
  Martin Do\v{s}k\'a\v{r}\inst{3}%\orcidID{0000-0001-7581-0567}
  \and
  Marc G.D. Geers\inst{1}%\orcidID{0000-0002-0009-6351}
  \and
  Vlado Menkovski\inst{4}%\orcidID{0000-0001-5262-0605}
  }
\institute{Department of Mechanical Engineering, Eindhoven University of Technology \and
DIFFER -- Dutch Institute for Fundamental Energy Research \and
Faculty of Civil Engineering, Department of Mechanics, Czech Technical University in Prague \and Department of Mathematics and Computer Science, Eindhoven University of Technology
\\
\email{f.hendriks@differ.nl, o.rokos@tue.nl, martin.doskar@cvut.cz, m.g.d.geers@tue.nl, v.menkovski@tue.nl}}

% You may leave out the orcidID information, if you want to.
% Use \corr to indicate the corresponding author. Note the spacing around the \corr command. Only one author can be the corresponding author.

%N.B.: comment out the \authorrunning{} command for the double-blind version of your paper submitted for review. Later, if your paper is accepted, use the command for the Camera-Ready Version.
% \authorrunning{A.L. Benjamin et al.}
% If there are two authors, write 'A.L. Benjamin and C.C. Broadus Jr.'
% If there are more than two authors, '[...] et al.' is used.

% \institute{Fictional Southern University, Savannah GA 31404, USA \email{\{a.l.benjamin,a.a.patton\}@fsu.fake}
% \and
% Fictional West Coast University, Long Beach CA 90840, USA \email{ccb@fwcu.fake}
% \and
% Secondary European Affiliation, Tiergartenstr. 17, 69121 Heidelberg, Germany
% \email{lncs@springer.com}

\maketitle

\begin{abstract}
Bifurcation phenomena in nonlinear dynamical systems often lead to multiple coexisting stable solutions, particularly in the presence of symmetry breaking. Deterministic machine learning models are unable to capture this multiplicity, averaging over solutions and failing to represent lower-symmetry outcomes. In this work, we formalize the use of generative AI, specifically flow matching, as a principled way to model the full probability distribution over bifurcation outcomes. Our approach builds on existing techniques by combining flow matching with equivariant architectures and an optimal-transport-based coupling mechanism. We generalize equivariant flow matching to a symmetric coupling strategy that aligns predicted and target outputs under group actions, allowing accurate learning in equivariant settings. We validate our approach on a range of systems, from simple conceptual systems to physical problems such as buckling beams and the Allen--Cahn equation. The results demonstrate that the approach accurately captures multimodal distributions and symmetry-breaking bifurcations. Moreover, our results demonstrate that flow matching significantly outperforms non-probabilistic and variational methods. This offers a principled and scalable solution for modeling multistability in high-dimensional systems.

\end{abstract}

\section{Introduction}
Machine learning enables efficient data-driven modeling of complex systems, especially when traditional methods impose high complexity or are incomplete \cite{senior2020improved,brunton2020machine,pfaff2020learning,hendriks2025similarity,minartz2025discovering,brinke2025flow}.
Many such systems exhibit bifurcations that mark sudden changes in system behavior due to a small change in a control parameter. Bifurcations are crucial in fields such as fluid dynamics \cite{ruelle1971nature}, climate science \cite{armstrong2022exceeding,willcock2023earlier,rietkerk2021evasion,steffen2018trajectories}, biology \cite{sadria2024fatenet}, and crowd dynamics \cite{muramatsu1999jamming,gu2025emergence,warren2022bifurcation}, where predicting transitions is essential.
Many bifurcating systems exhibit multistability, where multiple distinct stable states coexist under the same input parameters, and the number of these states can change as the parameters vary. This phenomenon often arises due to symmetry breaking, where a system's state after the bifurcation possesses fewer symmetries than before. This results in a multiplicity of possible solutions, all of which are equally valid.

For example, a fluid flow at high Reynolds number can start oscillating, breaking time-translation symmetry \cite{ruelle1971nature}. A symmetric structure under compression can buckle in multiple directions, breaking rotation and reflection symmetry \cite{bavzant2003stability}. An even mixture of phases can separate into distinct regions, breaking translational and reflection symmetry. Symmetry breaking is also central to magnetism, superconductivity, and particle physics \cite{beekman2019introduction,strocchi2005symmetry}, as well as the formation of enantiomers \cite{kondepudi2001chiral}. In mechanical metamaterials, buckling is a design feature that enables programmable shape transformations and tunable mechanical responses through controlled symmetry breaking \cite{kang2013buckling,azulay2024predicting}.

Simulating such systems with machine learning is nontrivial and remains an open problem \cite[\S 2.9.1]{zhang2025artificial}. Deterministic machine learning models can only produce the average of multiple solutions, resulting in nonphysical predictions that do not represent any true solution. Geometric deep learning approaches that rely on equivariant models preserve the system's symmetry but cannot select asymmetric outcomes, limiting their ability to capture bifurcations.
Approaches such as canonicalization \cite{lawrence2025improving}, input perturbation \cite{hendriks2025similarity}, model ensembles \cite{zou2025learning}, deterministic machine learning models combined with pseudo-arclength continuation \cite{fabiani2025enabling,della2025surrogate}, and sampling group operations to artificially break symmetry \cite{xie2024equivariant,lawrence2025improving} offer partial solutions but often lack generalizability or precision.

We propose modeling the full output distribution using generative modeling. Generative models aim to learn mappings from simple base distributions (e.g., Gaussian noise) to complex, high-dimensional target distributions. We use generative modeling to parameterize the distribution of possible bifurcation outcomes given the input parameters.

However, the probability distributions we require are singular: the set of allowed values (the support) lies on a subspace that has lower dimensionality than the input space, e.g., multiple Dirac deltas in 1D, or a circle in 2D space. This means the probability mass must be concentrated as sharply as possible around the allowed values.
When the target distribution is singular, the necessary mapping becomes highly nonlinear: nearby points in the source distribution may map to distant points in the target space. Capturing such mappings is challenging for neural networks, as it requires representing high-frequency functions \cite{rahaman2019spectral}. This is a fundamental limitation of direct generative approaches, which must learn a single, highly nonlinear transformation. For example, variational autoencoders (VAE) have a tendency to create `bridges' between modes, which is a common failure mode when modeling highly concentrated, multimodal distributions \cite{salmona2022can}. For this reason, they also tend to generate blurry images \cite{larsen2016autoencoding,bousquet2017optimal,dosovitskiy2016generating}.

In contrast, iterative methods such as diffusion models, denoising processes, and flow matching approximate this complex mapping as a sequence of small integration steps. This iterative structure distributes the required nonlinearity across many stages, making the learning problem more tractable, as each step only needs to model a small, smooth transformation. As a result, these methods are better suited for (i) modeling singular distributions, where the probability mass is concentrated on a low-dimensional manifold, and (ii) multimodal distributions, where the probability mass must split far apart \cite{salmona2022can}.

Therefore, to address the challenge of modeling multistability, we use flow  matching \cite{lipman2022flow} to train a stochastic interpolant \cite{albergo2022building} to model the full output distribution. We show that flow matching excels at approximating distributions whose probability mass is highly concentrated in a low-dimensional subspace (e.g., very close to two Dirac deltas). We further systematically integrate flow matching with equivariant modeling to tackle symmetry-breaking. Towards this end, we develop \emph{symmetric coupling}, an approach to determine the optimal training target, similar to minibatch optimal transport \cite{tong2023improving} and equivariant flow matching \cite{klein2023equivariant,song2023equivariant}, that exploits the self-similarity of the input. This approach substantially improves the training and the quality of generated flow paths. We demonstrate our method on abstract and physical systems, including coin flips, buckling beams, and phase separation via the Allen--Cahn equation. Our approach effectively captures bifurcations and multistability, with and without symmetry breaking.

The code is available at \url{https://github.com/FHendriks11/bifurcationML/}.

\section{Equivariance and Symmetry Breaking}

\paragraph{Equivariance} A map $f : \mathcal X \rightarrow \mathcal Y$ is equivariant with respect to a group $G$ if:
\begin{equation}\label{eq:equiv}
  g \cdot y = f(g \cdot x), \quad \forall x \in \mathcal X, \forall y \in \mathcal Y, \forall g \in G.
\end{equation}
If an input $x$ is invariant (i.e., self-similar) under a subset of the group actions (i.e., $g_x \cdot x = x$ for all $g_x \in G_x$, with $G_x \subseteq G$), then equivariance implies that output $y$ is also invariant (i.e., $g_x \cdot y = y$). However, this preservation of symmetry inherently prevents a model from expressing symmetry-breaking behavior. To capture bifurcations and transitions into asymmetric states, equivariant models must be adapted to allow for symmetry-breaking solutions, which means the traditional equivariance condition in Eq.~\eqref{eq:equiv} cannot hold for individual outputs after the bifurcation.

If there is symmetry-breaking in an equivariant system, there are multiple coexisting solutions for the same input. Meaning, if $y$ is a solution, then so are all $g_x \cdot y$, $\forall g_x \in G_x$ (i.e., the orbit of $y$ under $G_x$).

The equivariance in Eq.~\eqref{eq:equiv} then still holds for the set of solutions $\{ g_x \cdot y | g_x \in G_x\}$.
This is similar to the definition of ``relaxed equivariance'' by Kaba et al. \cite{kaba2023equivariance,kaba2023symmetry} in that both account for input symmetries via stabilizer subgroups (i.e., groups $G_x$ whose actions leave the input $x$ invariant), but it differs by preserving equivariance over sets of outputs (i.e., orbits of solutions), whereas relaxed equivariance modifies the equivariance condition itself, by allowing a correction with an element from the input's stabilizer subgroup.

\paragraph{As a probability distribution}
When using a generative model, we consider the set of solutions $\{y\}$ as a singular probability distribution $p(y|x)$, where the support of the distribution is the set of allowed solutions. The equivariance condition in Eq.~\eqref{eq:equiv} then translates to:
\begin{equation}\label{eq:equiv_pdf}
  p(y|x) = p(g\cdot y | g \cdot x) \quad \forall x \in \mathcal X, y \in \mathcal Y, g \in G.
\end{equation}

To ensure that the predictions by the model satisfy Eq.~\eqref{eq:equiv_pdf}, we use (i) a model that is equivariant under $G$ to parametrize the probability distribution and (ii) a prior $p(y_0)$ that is invariant under $G$. This ensures that the generated probability distribution respects the symmetry of the problem \cite{minartz2023equivariant}.

\paragraph{Flow matching} To model this probability distribution, we use flow matching \cite{lipman2022flow}, which learns a vector field $u(y_t, t, x)$ that transforms samples from an uninformed prior $p(y_0)$ into samples $y_1$ from the target distribution $p(y|x)$ over a continuous pseudo-time variable $t \in [0, 1]$. The vector field is learned by minimizing the expected squared error between the model prediction and the target vector field along linear interpolation paths between prior and target samples.

\paragraph{Symmetric coupling}
Additionally, we leverage the fact that there are multiple equivalent outputs to improve our predictions, using \emph{symmetric coupling}.
During training, for each latent sample $y_0$ and target $y_1$, we find the closest equivalent under $G_x$ to the current prediction, i.e., find
\begin{align}\label{eq:symmetriccoupling}
  y_1' &= \tilde{g}_x \cdot y_1 \\
  \tilde{g}_x &= \text{argmin}_{g_x \in G_x} c(y_0, g_x \cdot y_1),
\end{align}
where $c$ is the cost function, for which the Euclidean distance squared $c = ||y_0 - y_1||^2$ is a possible choice.

This is a way of straightening the flow paths similar to minibatch optimal transport \cite{tong2023improving}, but instead of finding the optimal coupling between a minibatch of the prior samples and target samples, we find the optimal coupling between one prior sample $y_0$ and all symmetric equivalents $g_x\cdot y_1, g_x \in G_x$ of its corresponding target sample $y_1$. Our approach is similar to equivariant flow matching as described by Klein \cite{klein2023equivariant} and Song \cite{song2023equivariant}, except we only consider the group $G_x$ that leaves the input $x$ invariant. If this group contains permutations or rotations, we can therefore follow their lead and use the Hungarian algorithm for permutations or the Kabsch algorithm for rotations. Other options are checking all possible reflections of the target or using FFT cross-correlation to determine the optimal periodic translation of the target.

\begin{figure*}
  \centering
  \includegraphics[width=\textwidth]{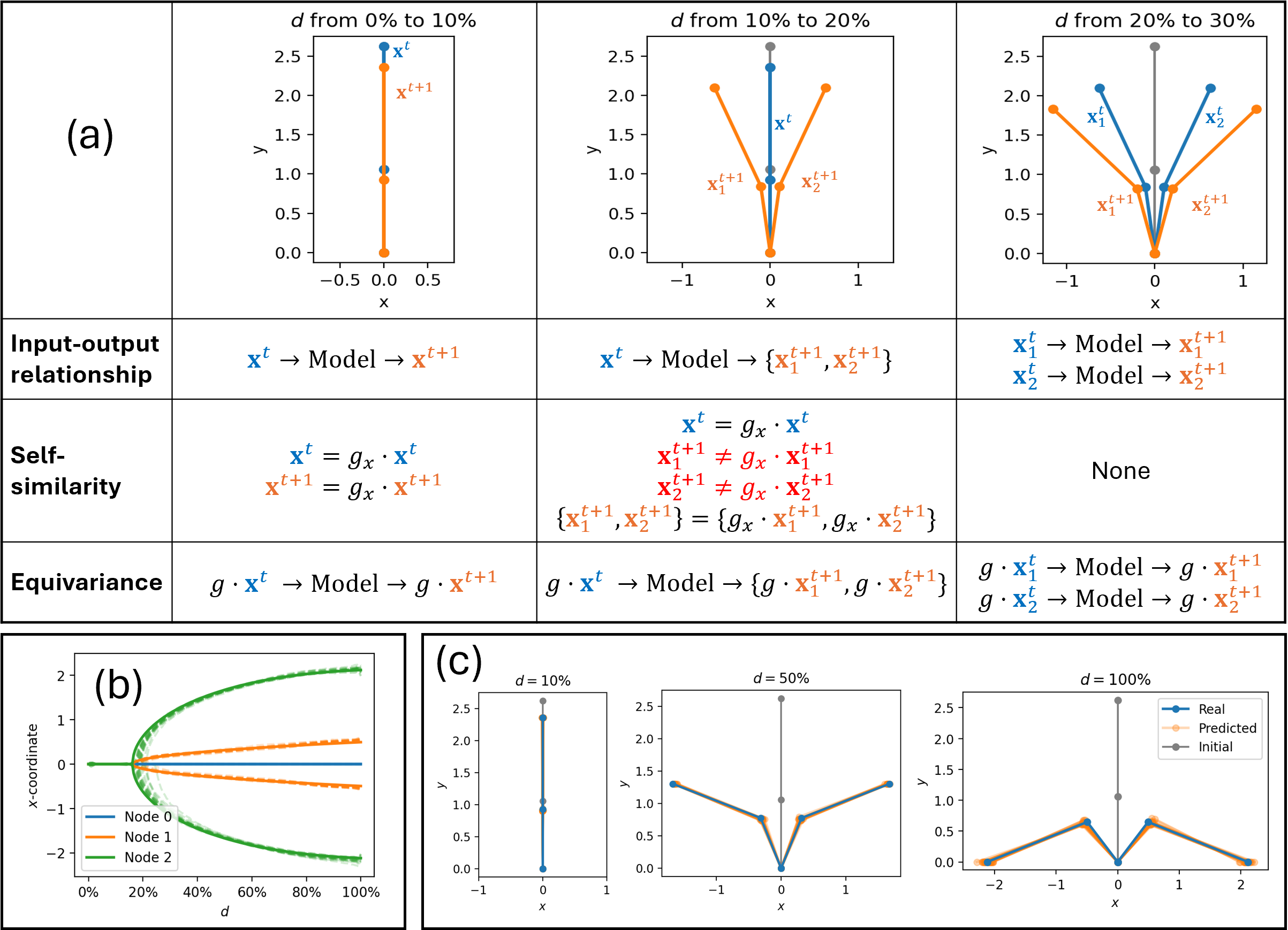}
  \caption{(a) Self-similarity and equivariance illustrated with the buckling beam problem, in which the input is the state at the current time step $t$ and the target output is the next time step $t+1$. $d$ is the vertical downward displacement of the beam tip relative to the total length of the beam, which is increased over time. The relevant symmetry group $G$ here consists of only two elements: identity and reflection in the y-axis. In this case, the self-similarity of the beam before buckling happens to be described by the same group, i.e., here $G_x=G$. (b) The $x$-coordinates of each node over the course of a trajectory, showing 50 predictions of a trained flow matching model (dashed lines) compared to the two possible ground truths (solid lines). (c) The same 50 predictions compared to the two possible ground truths, showing the beam deformation.}
  \label{fig:equivariance}
\end{figure*}

See Fig.~\ref{fig:equivariance}(a) for a visualization of the concepts of equivariance and self-similarity in the context of a buckling beam problem.

\section{Experiments and Results}

Here, we showcase the performance of the proposed approach with four toy problems and two physically grounded examples.

All models were trained with the Adam optimizer.

% \subsection{Datasets}
% \label{app:datasets}

\subsection{Gaussian to 2 Dirac Deltas}
This toy problem demonstrates the ability of flow matching to capture multimodal distributions with very sharp peaks. There is no fixed dataset size because, during training, we continuously sample new data points. The input $x_i$ is sampled from a Gaussian distribution $\mathcal{N}(0, 1)$; the output $y_i$ is sampled from $\{-1, 1\}$ with equal probability, or equivalently from $p(y_i) = 0.5 \delta(y_i - 1) + 0.5 \delta(y_i + 1)$.

For the VAE, we used an MLP with 2 hidden layers of 64 neurons per layer as an encoder and another MLP as a decoder, with a latent dimension of 16. For the flow model, we used an MLP with 3 hidden layers and 64 neurons per layer. Both models were trained for 10000 epochs with a batch size of 256. We used a learning rate of $10^{-3}$ for flow matching and $10^{-2}$ for the VAE.

\begin{figure}[!htbp]
  \centering
  \includegraphics[width=0.7\linewidth]{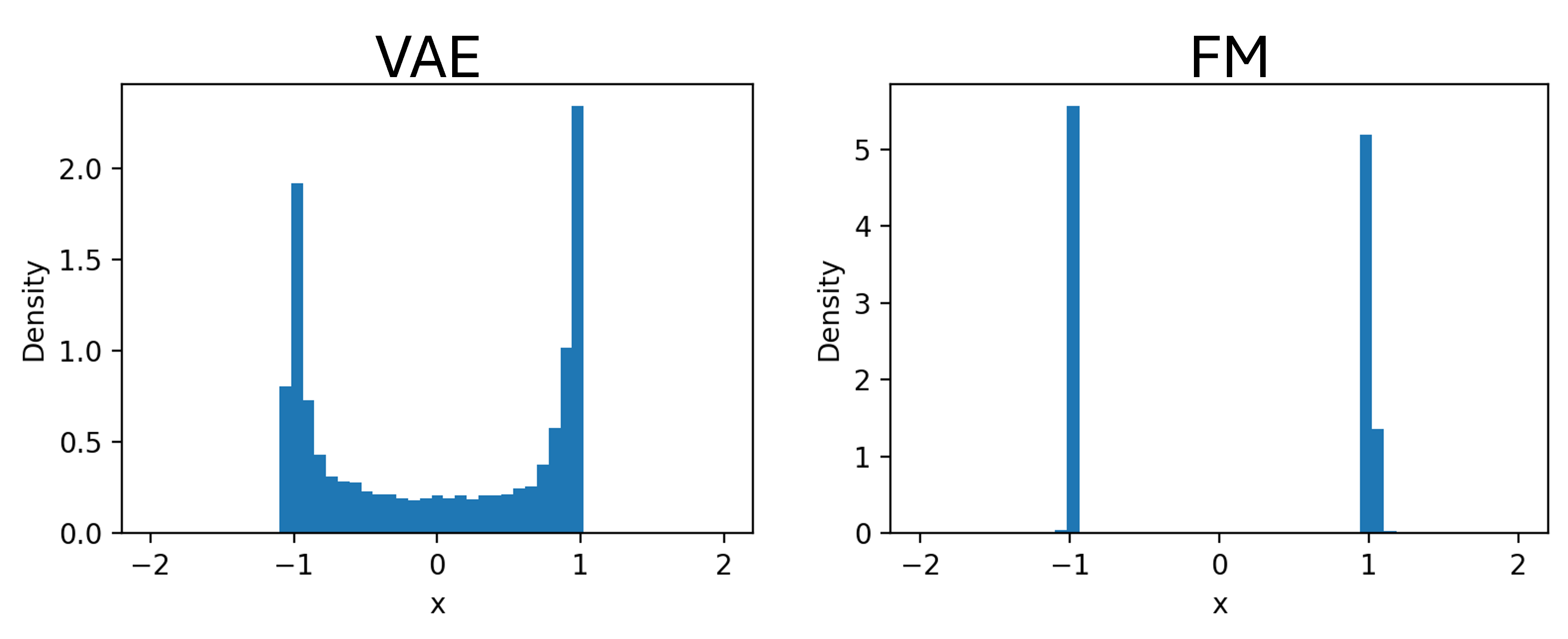}
  \includegraphics[width=0.8\linewidth]{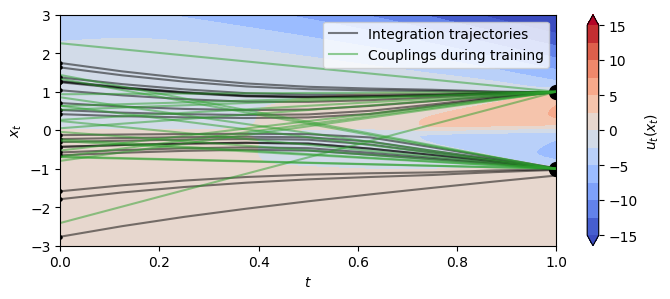}
  \vspace{0.5em}
  \includegraphics[width=0.8\linewidth]{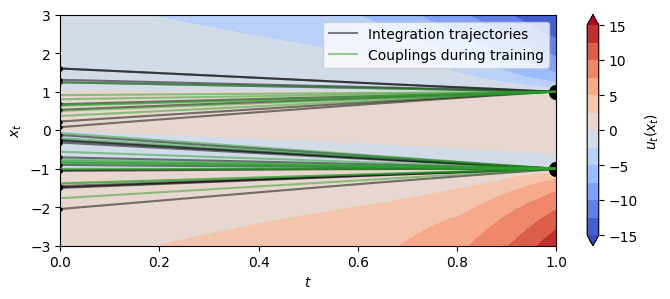}
  \caption{Top: the results of predicting a probability distribution consisting of two Dirac deltas, using a VAE and flow matching. Middle/bottom: the learned flow field by the trained flow matching model (contour plot), showing with green lines the coupling during training between prior and targets (middle: without symmetric matching, bottom: with) and how the coupling samples from the Gaussian prior are pushed towards the two delta peaks (black lines). \label{fig:gaussian_to_dirac}}
\end{figure}

Figure~\ref{fig:gaussian_to_dirac} compares predictions from the VAE and flow matching, showing that the VAE gives a less sharp distribution and creates a `bridge' between the solutions, while flow matching gives a much sharper distribution that is more concentrated around the two peaks.
The figure also visualizes the learned flow field with and without symmetric coupling, illustrating how symmetric coupling straightens the flow paths.

% \FloatBarrier

\subsection{Coin Flip}
To model a coin flip, we take as input the amount of money one bets (positive for heads, negative for tails), and then predict the winnings as output. This means the outcome is either the same as the input (winning the bet amount) or the negative of the input (losing the bet amount). Our dataset consists of 1000 inputs $x_i$ sampled uniformly from the range $[-100, 100]$, with corresponding outputs $y_i \in \{x_i, -x_i\}$ with equal probability. We use 800 of these data points for training and 200 for testing.

As a non-probabilistic baseline, we used an MLP with 2 hidden layers of 32 neurons per layer, trained for 52 epochs (early stopping), with batch size 64 and learning rate $10^{-4}$. For the VAE, we used an MLP encoder-decoder with 2 hidden layers of 32 neurons, latent dimension 16, and an additional MLP for conditioning on the bet amount. The VAE was trained for 1000 epochs with learning rate $10^{-3}$. For flow matching, we used a similar MLP to predict the flow field with Gaussian prior noise of standard deviation 1.0.

Even though the model is not equivariant, we still test symmetric coupling in the form of picking $-1$ or $+1$ depending on which is closer to the prior sample.

\begin{figure*}[!htbp]
  \centering
  \includegraphics[width=0.90\textwidth]{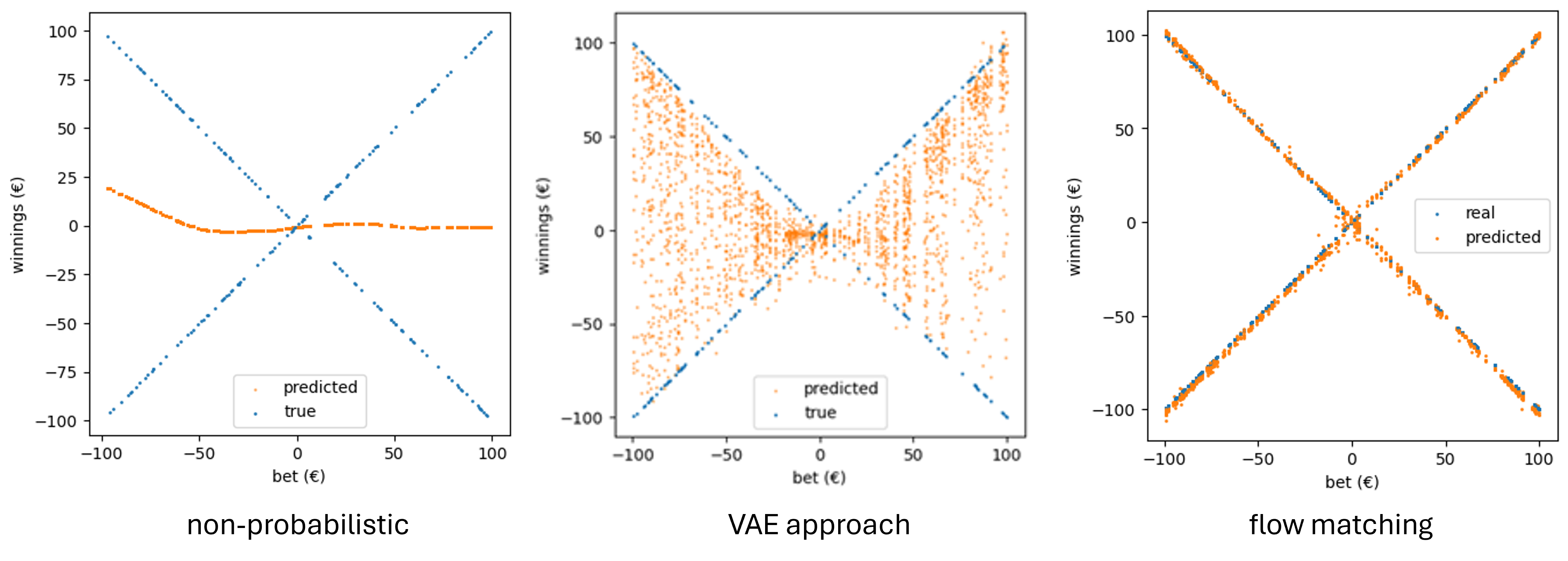}
  \caption{The results of predicting outcomes of a coin flip using three different methods: (a) a regular neural network, (b) a conditional VAE, and (c) flow matching. Each method makes 10 predictions per test data point. \label{fig:coinflip}}
\end{figure*}

Figure~\ref{fig:coinflip} compares deterministic, VAE, and flow matching predictions, showing that the deterministic model fails to capture the bimodality, while the VAE captures it but with a blurry distribution that is not concentrated around the two solutions. In contrast, the flow matching model captures the bimodality with a much sharper distribution.

\subsection{Three Roads}
\begin{figure}[!htbp]
  \centering
  \includegraphics[width=0.9\linewidth]{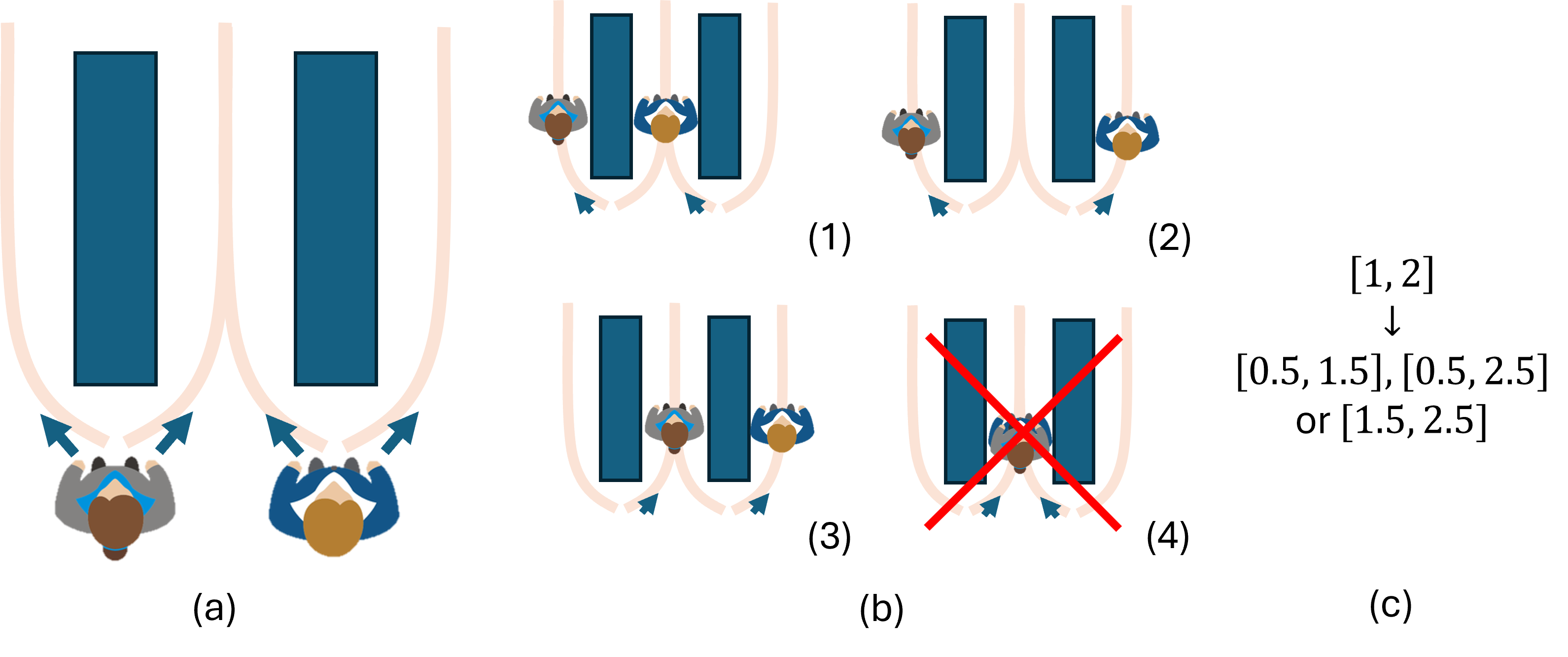}
  \caption{Illustration of the 3 roads problem. (a) Input, (b) possible outputs, (c) what that looks like in the actual data. \label{fig:threeroads}}
\end{figure}

This test system models bifurcation in a situation where there are two entities (can be nodes in a graph or something else) that need to coordinate.
One can imagine that these are two people that are in a store with very narrow aisles. Both of them are standing right in front of an obstacle separating the aisles, and want to avoid this obstacle by either swerving left or right. However, they don't want to bump into each other, so if one of them chooses the middle aisle (e.g., going left for the right person, going right for the left person), the other will not pick the same aisle. See Fig.~\ref{fig:threeroads} for an illustration. This is important in prediction of, e.g., pedestrian dynamics, where 1) people coordinate to avoid bumping into each other, and 2) the distributions of their possible paths are highly multimodal when obstacles are involved; there are multiple ways to go around an obstacle, but none to go through it \cite{minartz2025discovering,brinke2025flow}.
We choose to model this as a very simple system with two input features and two output features. The input and output features $[x_i^{(1)}, x_i^{(2)}]$ and $[y_i^{(1)}, y_i^{(2)}]$ are two numbers that indicate the horizontal position of both entities before and after deciding their path, respectively. The possible values of $[y_i^{(1)}, y_i^{(2)}]$ are
\begin{align}
  \begin{split}
    [y_i^{(1)}, y_i^{(2)}] \in & \begin{cases}
      [x_i^{(1)} - d/2, x_i^{(2)} + d/2]\\
      [x_i^{(1)} - d/2, x_i^{(2)} - d/2]\\
      [x_i^{(1)} + d/2, x_i^{(2)} + d/2]
    \end{cases}\\
    \text{with } d = x_i^{(2)} - x_i^{(1)}.
  \end{split}
\end{align}
For example, for the input $[x_i^{(1)}, x_i^{(2)}]=[1.0, 2.0]$, the output could be $[0.5, 2.5]$, $[0.5, 1.5]$, or $[1.5, 2.5]$.
In this test system, we have chosen to consider all these possibilities equally likely. Our dataset consists of 2000 data points, with $[x_i^{(1)}, x_i^{(2)}]$ sampled uniformly from the range $[-50, 50]$. Of the 2000 data points, 1600 are used for training and 400 for testing.

We built all approaches (non-probabilistic, VAE, flow matching) on a set model that applies MLPs to each element and each ordered pair of elements, with permutation-equivariant aggregation. As a prior for the flow matching model, we used Gaussian noise with standard deviation 1.0.

\FloatBarrier

\subsection{Four Node Graph}
\begin{figure}[!htbp]
  \centering
  \includegraphics[width=0.4\linewidth]{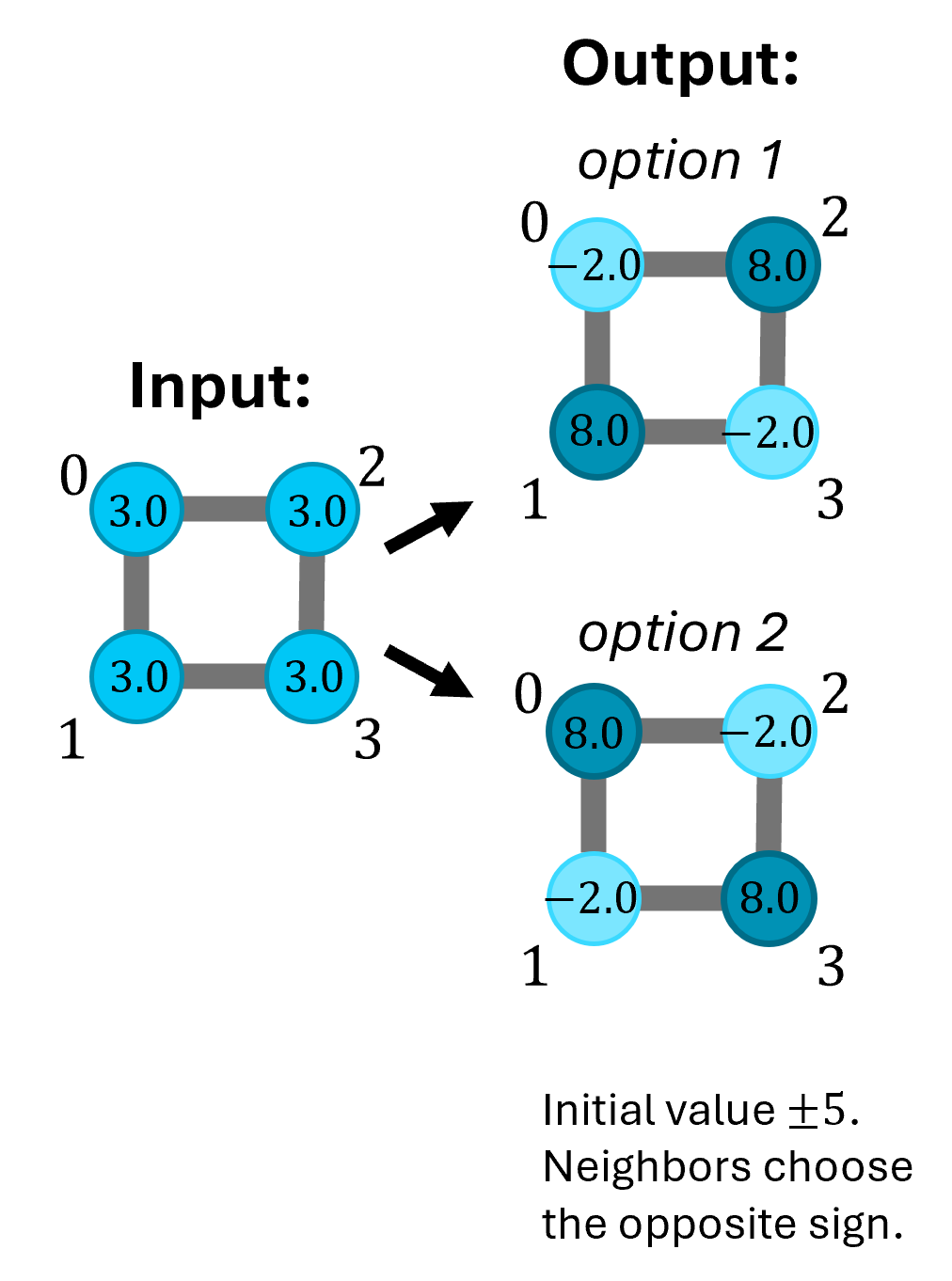}
  \caption{Illustration of the four node graph problem. \label{fig:fournodegraph}}
\end{figure}

For this problem, the input is a graph consisting of 4 nodes, connected in a square. Each node has the same value $x_i$, as a node embedding, which is sampled uniformly from the range $[-50, 50]$. The output is an embedding for each node which is equal to $x_i \pm 5$, with connected nodes choosing the opposite sign.
See Fig.~\ref{fig:fournodegraph} for a visualization. This system is equivariant with respect to permutation. Additionally, the input is self-similar under some permutations, whereas each individual output is self-similar under fewer permutations, which means there is symmetry-breaking.
Our dataset contains 2000 data points, of which 70\% is used for training, 30\% for testing.

As a non-probabilistic model, we used a message-passing graph neural network with 3 message-passing layers. For the VAE, we used such a message-passing GNN as encoder and another as decoder, and another to condition on the input graph. For flow matching, we used a similar message-passing GNN with the input value plus Gaussian noise (standard deviation 1.0) as a prior. Symmetric coupling is implemented by choosing the target permutation with minimum Euclidean distance squared to the noise sample. Only two permutations are relevant here: identity, and the swap of nodes 0 with 2 and 1 with 3.

% \FloatBarrier

\subsection{Buckling Beam}
For this problem, we consider a beam that is attached to a base on the bottom, and then prescribe a vertical displacement to the top of the beam, while leaving the horizontal displacement free. For a small displacement, the beam will compress but stay straight, but when the displacement reaches a critical value, the system becomes unstable and the beam will buckle, which will make it bend to either the left or right side, with either option equally likely.
We model the beam as a series of $n$ connected segments, which connect $n+1$ nodes, where node 0 is attached to the base. We consider $n$ ranging from 2 to 10. Each segment has a length $L_i$ and an axial stiffness $C_i$. Individual segments can rotate relative to each other, reflected by a rotational stiffness $K_i$. The input $d$ is the vertical displacement applied to the top node.
We solve the deformation of the beam by finding the Hencky strains $\epsilon_i$ and angles $q_i$ that minimize the total energy:
\begin{equation}
U(\mathbf{q}, \epsilon) = \frac{1}{2}\left[K_1q_1^2 + \sum_{i=2}^n K_i(q_i-q_{i-1})^2+ \sum_{i=1}^n L_i C_i \epsilon_i^2\right]
\end{equation}
subject to the constraint that the top of the beam is at the right displacement $d$:
\begin{equation}
\sum_{i=1}^n L_i \left[1-\exp(\epsilon_i)\cos q_i\right] = d.
\end{equation}
We solve this using Lagrange multipliers, combined with stability analysis. The solution is perturbed using the lowest eigenmode of the stiffness matrix to break symmetry when the system becomes unstable.
In total, we generate 1000 beams. We sample $n$ from the integers 2 to 11, and sample $L_i$, $C_i$ and $K_i$ from the log-uniform distribution between 0.5 and 2. For each beam we solve for its trajectory as $d$ is increased from zero. We choose the range of $d$ such that at the end of a trajectory the top node is at the ground, and we divide this range into 200 time steps.
70\% of the resulting trajectories are used for training, 30\% for testing.

For this system, we used a flow matching model with a random walk prior similar to \cite{brinke2025flow}, with an EGNN \cite{satorras2021n} to predict interactions between nodes and UNets for time evolution. Symmetric coupling selects the closest reflected target trajectory.

Figure~\ref{fig:equivariance}(b,c) shows representative predictions against both valid ground truths, showing that the model can accurately capture both solution branches.

\subsection{Allen--Cahn}
The Allen--Cahn equation is a reaction-diffusion equation and describes the time evolution of a binary mixture of phases, e.g., ice and water at zero degrees Celsius. It is given by
\begin{equation}
  \frac{\partial u}{\partial t} = \epsilon^2 \frac{\partial^2 u}{\partial x^2} - (u^3 - \mu u),
\end{equation}
where $u$ is the phase-field variable, representing the concentration of one of the phases. $\epsilon$ is a parameter that controls the interface width between phases, and $u^3 - \mu u$ is the derivative of a double-well potential with minima at $\pm \sqrt{\mu}$ if $\mu > 0$; for $\mu \leq 0$, there is only one minimum at $0$.
We solve this equation on a 1D domain of length 1.0 with periodic boundary conditions, from $t=0$ to $t=100$, using a finite difference scheme with 200 spatial points and a time step of 0.1. Each trajectory starts from the homogeneous state $u = 0$, however, because this is an unstable state when $\mu>0$, we add a small random perturbation sampled from $\mathcal{N}(0, 0.001)$ to each spatial point. We generate 400 trajectories in total, of which 300 are used for training and 100 for testing. For each trajectory, $\epsilon$ is randomly sampled from a log-uniform distribution between 0.001 and 0.1, and $\mu$ from $\mathcal{U}(-0.1, 1.0)$.

We used a random walk prior (cumulative scaled Gaussian noise) and a 2D UNet to predict the flow field from the entire random-walk trajectory to the entire final trajectory, with periodicity only in the spatial dimension. The UNet has 4 downsampling steps and 32 channels. Symmetric coupling combines the closest circular shift (via FFT cross-correlation), reflection, and sign flip, which improves performance and enables recovery of the pitchfork bifurcation behavior.

\begin{figure}[!htbp]
  \centering
  \includegraphics[width=0.75\linewidth]{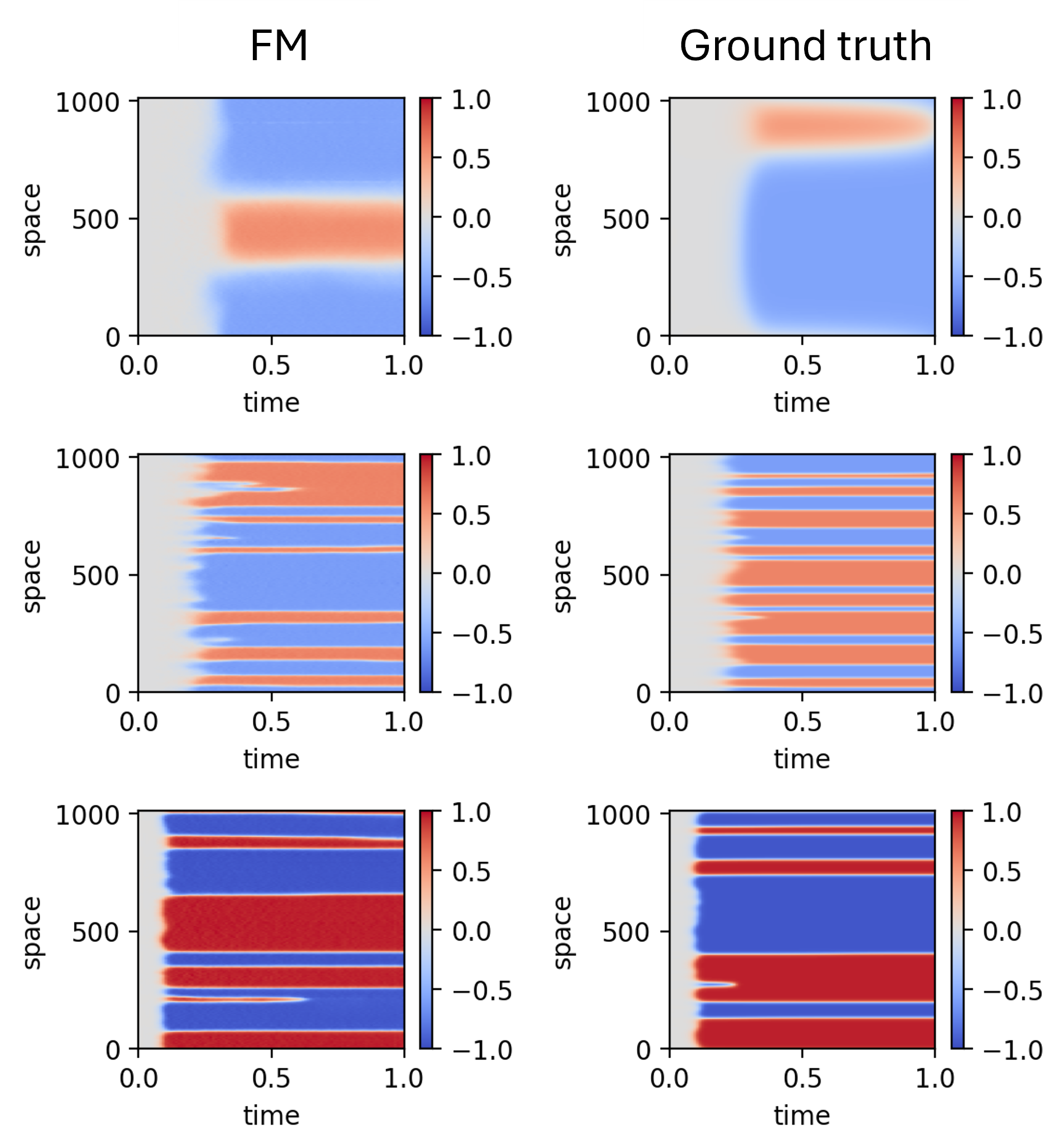}
  \caption{Comparison of predicted Allen--Cahn trajectories by the flow model with symmetric coupling with examples of possible ground truth trajectories. \label{fig:allen_cahn_trajectories}}
\end{figure}

\begin{figure}[!htbp]
  \centering
  \includegraphics[width=0.49\linewidth]{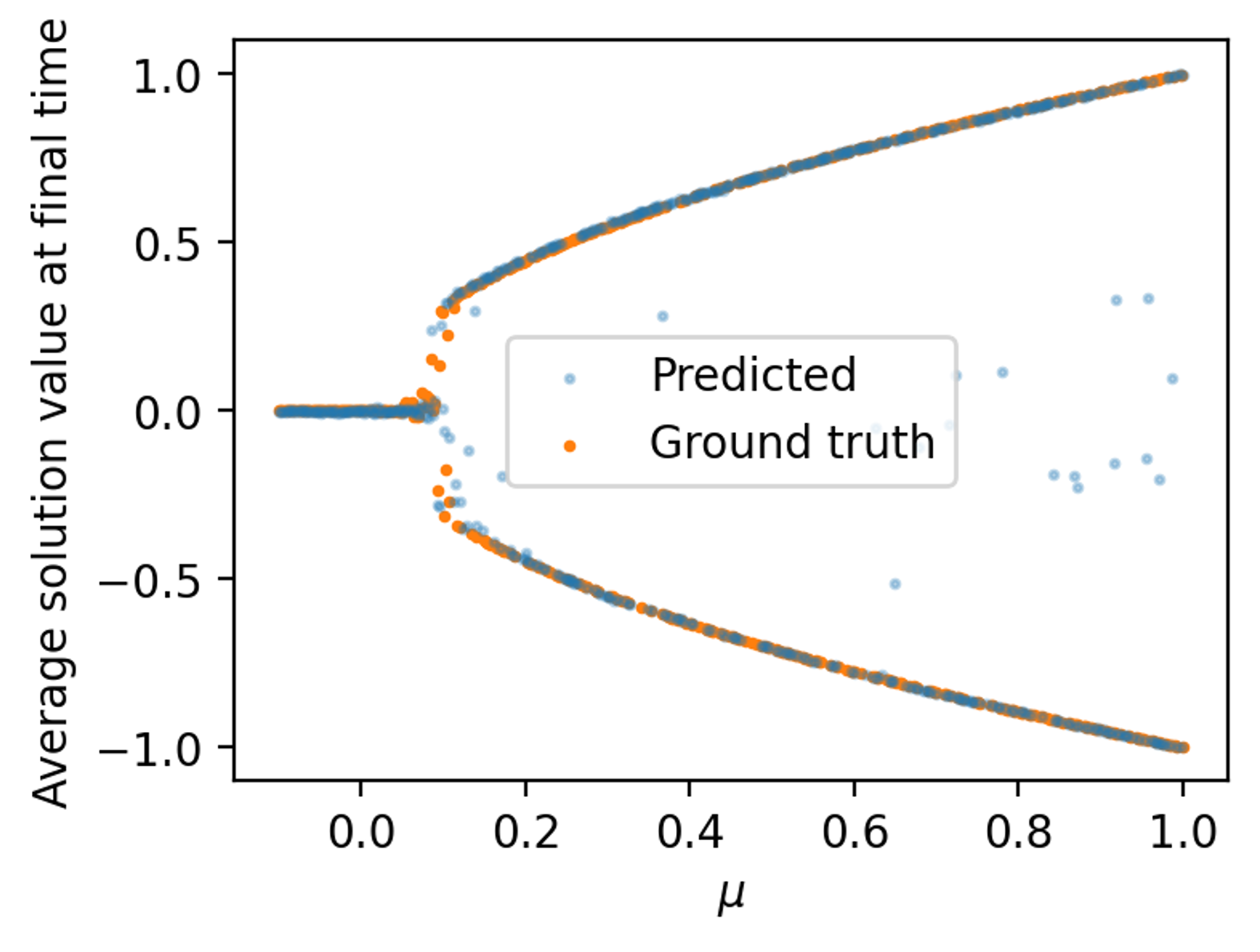}
  \includegraphics[width=0.49\linewidth]{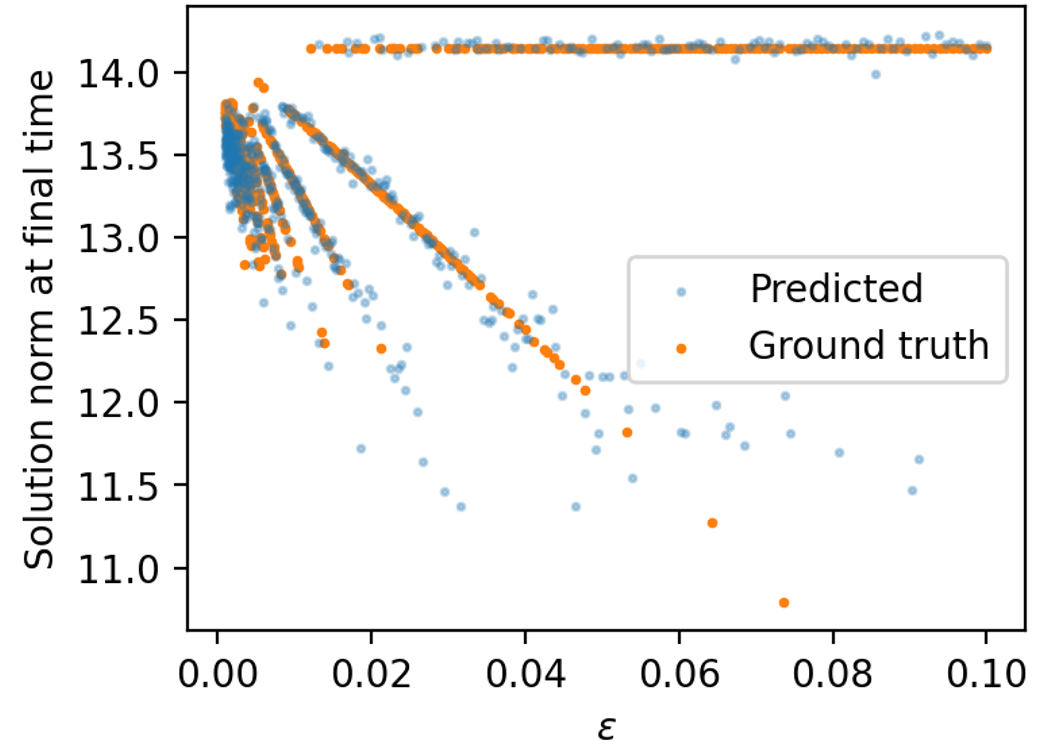}
  \caption{Bifurcation diagrams of the Allen--Cahn equation. Left: the parameter $\mu$ is varied, keeping $\epsilon$ constant at 0.1. The trained model is able to model the pitchfork bifurcation. Right: the parameter $\epsilon$ is varied, keeping $\mu$ constant at 1.0. The model is able to capture the addition of new stable states as $\epsilon$ decreases.\label{fig:allen_cahn_bifurcation}}
\end{figure}

See Figure \ref{fig:allen_cahn_trajectories} for several predictions by the trained flow model, showing that even though the solutions are not the same as the ground truth, they do reproduce the same qualitative behavior. This is also illustrated by Figure \ref{fig:allen_cahn_bifurcation}, which shows that the trained model can reproduce bifurcation diagrams of the system.

% \FloatBarrier

\subsection{Overall quantitative comparison}
Table~\ref{tab:wasserstein} reports the mean Wasserstein distance between the predicted and actual distribution of allowed outcomes for each test data point (except for Allen--Cahn, where we report the governing equation residual). For the 2-Delta Peaks dataset and the 4-node graph, we also compare to symmetric coupling. In all cases, the flow matching outperforms the non-probabilistic model and the VAE, and symmetric coupling further improves the result. For the 2-Delta Peaks, even though the model is not equivariant, symmetric coupling still improves performance because the targets and the prior are symmetric to a sign flip.

\begin{table}
  \caption{Performance of various approaches for the different test systems. All metrics except for the Allen--Cahn are the Wasserstein distance of a set of 100 predictions to the actual distribution of allowed outcomes for each test data point, averaged over all test data points. For the Allen--Cahn system, we use the residual of the governing equation as a performance metric, acknowledging its limitations due to sensitivity to small fluctuations in the prediction, particularly in the Laplacian term.\label{tab:wasserstein}}
  \centering
  \begin{tabular}{lllll}
    \toprule
    Test System     & Non-prob. & VAE & FM  & FM*  \\
    \midrule
    Two Delta Peaks & 1.0**                      & 0.25          & 0.091     & \textbf{0.0041}        \\
    Heads or Tails  & 56.2                       & 33.0         & \textbf{8.30}         & -          \\
    Three Roads     & 21.4                       & 17.3         & \textbf{3.88}           & -         \\
    Four Node Graph & 10.0                       & 9.89         & 2.02  & \textbf{1.19}           \\
    Buckling Beam & -                       & -         & 23.1 & \textbf{9.6}           \\
    Allen--Cahn & -                       & -      & 255   & \textbf{244} \\
    \bottomrule
    \multicolumn{5}{p{0.85\linewidth}}{\small{*With symmetric coupling.}}\\
    \multicolumn{5}{p{0.85\linewidth}}{\small{**Theoretical result, a non-probabilistic model would always predict the mean 0.}}
  \end{tabular}
\end{table}

\FloatBarrier

\section{Conclusion}
\label{sec:conclusions}
We introduced a probabilistic, equivariant flow matching framework to address the challenge of modeling multiple coexisting solutions in symmetry-breaking bifurcation problems. By leveraging generative modeling and symmetric coupling, our approach captures multimodal output distributions while respecting system symmetries. Through experiments on both abstract and physically grounded systems, we demonstrated that our method outperforms traditional and variational models in accurately representing bifurcation behavior. This work opens new avenues for integrating symmetry-aware generative models into the analysis of complex dynamical systems.

\begin{credits}
\subsubsection{\ackname}
This project has received funding from the Eindhoven Artificial Intelligence Institute (EAISI). MD's work was co-funded by the European Union under the project Robotics and Advanced Industrial Production (reg. no. CZ.02.01.01/00/22\_008/0004590).

%  A bold run-in heading in small font size at the end of the paper is
% used for general acknowledgments, for example: This study was funded

\subsubsection{\discintname}
The authors have no competing interests to declare that are relevant to the content of this article.
% It is now necessary to declare any competing interests or to specifically
% state that the authors have no competing interests. Please place the
% statement with a bold run-in heading in small font size beneath the
% (optional) acknowledgments,
% for example: The authors have no competing interests to declare that are
% relevant to the content of this article. Or: Author A has received research
% grants from Company W. Author B has received a speaker honorarium from
% Company X and owns stock in Company Y. Author C is a member of committee Z.
\end{credits}

\bibliographystyle{splncs04}
\bibliography{library}

\end{document}